\documentclass[10pt,twocolumn,letterpaper]{article}

\usepackage[accsupp]{axessibility}

\usepackage{iccv}              
\usepackage{placeins}

\newcommand{\red}[1]{{\color{red}#1}}
\newcommand{\todo}[1]{{\color{red}#1}}
\newcommand{\TODO}[1]{\textbf{\color{red}[TODO: #1]}}

\definecolor{iccvblue}{rgb}{0.21,0.49,0.74}
\usepackage[pagebackref,breaklinks,colorlinks,allcolors=iccvblue]{hyperref}

\def\paperID{7} 
\def\confName{ICCV Workshop P13N}
\def\confYear{2025}

\title{MINDiff: Mask-Integrated Negative Attention\\for Controlling Overfitting in Text-to-Image Personalization}

\author{Seulgi Jeong \quad Jaeil Kim\textsuperscript{*}\\
Kyungpook National University\\
Daegu, South Korea\\
{\tt\small \{wjdtmfrl3682,  jaeilkim\}@knu.ac.kr}
}

\usepackage{booktabs}
\usepackage{pifont}
\usepackage[table]{xcolor}

\begin{document}
\maketitle

\begingroup
\renewcommand\thefootnote{}
\footnotetext{\textsuperscript{*}Corresponding author.}
\endgroup

\begin{abstract}
In the personalization process of large-scale text-to-image models, overfitting often occurs when learning specific subject from a limited number of images. Existing methods, such as DreamBooth, mitigate this issue through a class-specific prior-preservation loss, which requires increased computational cost during training and limits user control during inference time. To address these limitations, we propose \textbf{Mask-Integrated Negative Attention Diffusion (MINDiff)}. MINDiff introduces a novel concept, \textbf{negative attention}, which suppresses the subject’s influence in masked irrelevant regions. We achieve this by modifying the cross-attention mechanism during inference. This enables semantic control and improves text alignment by reducing subject dominance in irrelevant regions. Additionally, during the inference time, users can adjust a scale parameter $\lambda$ to balance subject fidelity and text alignment. Our qualitative and quantitative experiments on DreamBooth models demonstrate that MINDiff mitigates overfitting more effectively than class-specific prior-preservation loss. As our method operates entirely at inference time and does not alter the model architecture, it can be directly applied to existing DreamBooth models without re-training. Our code is available at \url{https://github.com/seuleepy/MINDiff}.
\end{abstract}

\section{Introduction}
\label{sec:introduction}

\begin{figure*}
    \centering
\includegraphics[width=1.0\linewidth]{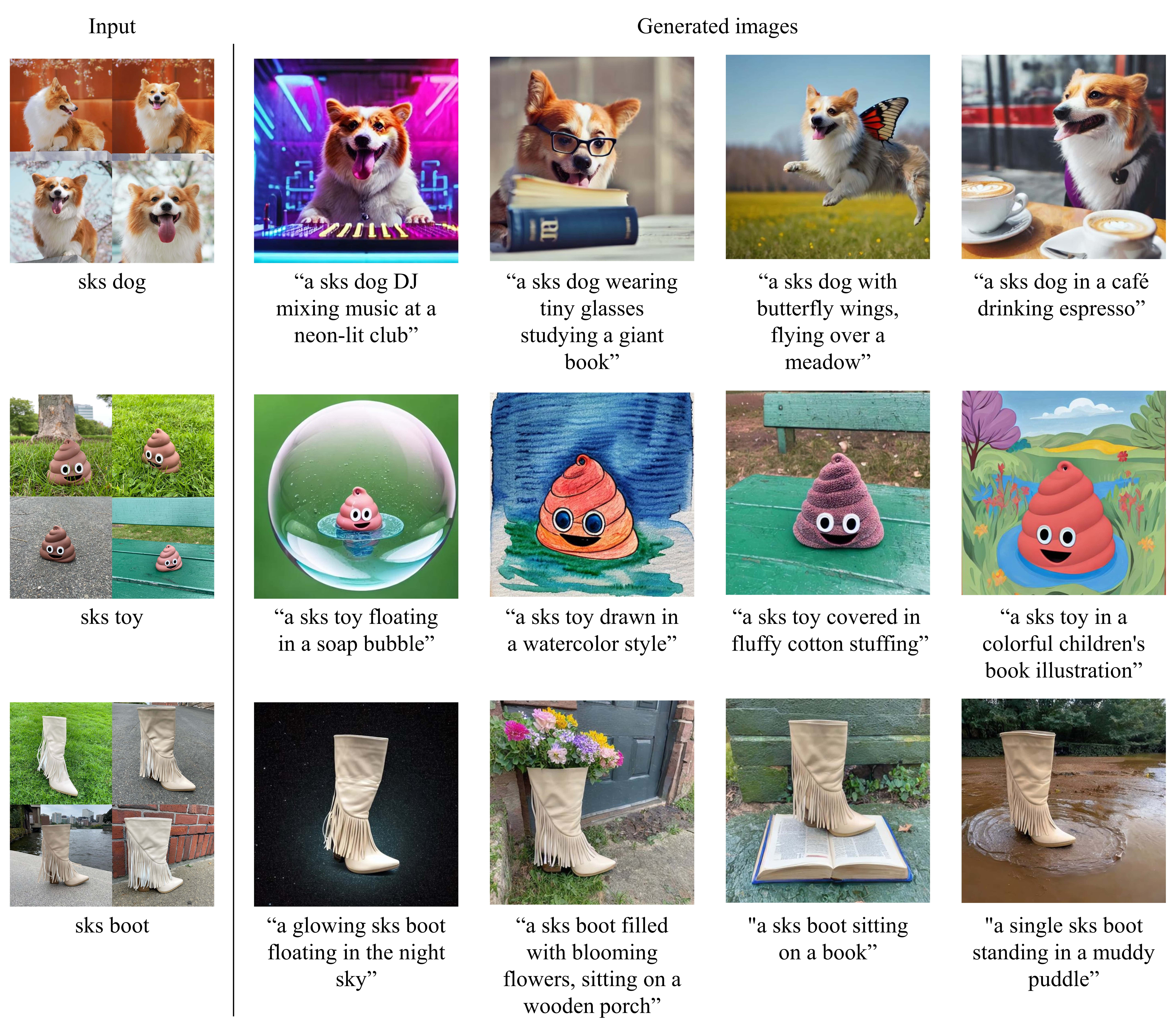}
    \caption{Qualitative results of applying MINDiff to the DreamBooth model. The first column shows input images and the remaining columns present images generated from various text prompts, demonstrating the model’s ability to preserve subject identity while adapting to textual descriptions. All results are based on Stable Diffusion 1.4.}
    \label{fig:mindiff_result}
\end{figure*}

Recent advances in large-scale text-to-image generation \cite{nichol2021glide, parti, ldm, dalle, dalle2, imagen} models have enabled the creation of novel scenes from textual descriptions and the localized image edits. However, these methods still face challenges in preserving the identity of specific subjects while generating diverse images across different contexts. For example, when generating images of an elderly individual with mobility impairments in wedding or travel scenarios, it is essential to preserve their distinctive features while appropriately modifying the background and context.

Regarding these limitations, personalization techniques have been explored to enable generative models recognize and incorporate user-defined subjects. One of the principal approaches is DreamBooth \cite{dreambooth}, which fine-tunes a pre-trained text-to-image model using a small set of sample images to capture the unique characteristics of the subject. DreamBooth is prone to overfitting; to address this, it introduces a class-specific prior preservation loss (PPL). However, this strategy increases training cost and still struggles to balance text alignment and subject fidelity. Furthermore, it can lead to cases where, instead of the intended subject, another object of the same class appears in the generated image.

In addition to ensuring subject fidelity and text alignment, it is important to provide controllability, defined here as the ability to adjust the trade-off between these two competing objectives.
However, existing methods frequently struggle to maintain this balance consistently. Their performance tends to vary depending on factors such as the initial latent code, the given prompt, or the characteristics of the learned concept.
These inconsistencies highlight the need for intuitive user control, where adjustable parameters can enhance both reliability and flexibility in personalized image generation.

In this paper, we propose a novel inference-time method to mitigate overfitting in personalized text-to-image generation.
MINDiff introduces \textit{negative attention}, a novel concept that suppresses subject influence in spatially masked regions via a modified cross-attention mechanism.
Unlike prior-preservation loss, which introduces additional training cost to mitigate overfitting, our method achieves the same goal without incurring such overhead.
Furthermore, it allows users to control the trade-off between subject fidelity and text alignment through a tunable scale parameter $\lambda$ during inference.
As it operates entirely at inference time, it can be applied directly to existing DreamBooth models without modifying or re-training the model weights.
\Cref{fig:mindiff_result} shows results of applying MINDiff to a DreamBooth model. 
These results demonstrate that our method preserves subject identity, improves text alignment, thereby enabling the generation of diverse outputs across prompts.

Our key contributions are as follows:
\begin{enumerate}
    \item We propose a novel concept, negative attention, which suppresses the subject’s influence in masked irrelevant regions within the cross-attention mechanism.
    \item This approach mitigates overfitting in DreamBooth models without requiring class-specific prior-preservation loss.
    \item We enable users to control the trade-off between subject fidelity and text alignment at inference time via a tunable scale parameter, $\lambda$.
    \item Since our method operates entirely at inference time, it can be directly applied to models that have been fine-tuned with DreamBooth, without modifying their weights.
\end{enumerate}

\section{Related Work}
\label{sec:related_work}

\subsection{Text-to-Image Generation}
\label{sec:text-to-image_generation}
Text-to-image generation has evolved through various generative modeling approaches, including GANs~\cite{2016gan, stackgan, controlgan, dmgan, vqgan, xu2018attngan, dfgan}, autoregressive models~\cite{parti, dalle, make-a-scene, ding2021cogview}, and diffusion models~\cite{nichol2021glide, dalle2, imagen, ldm}.
Diffusion models generate images through iterative denoising.
They have demonstrated strong performance in terms of text-image alignment and visual quality~\cite{diffusion_beat}.

\subsection{Mask-Based Image Editing}
\label{sec:mask_based_image_editing}

Recent diffusion-based image editing methods commonly adopt masking strategies for localized editing.
While MINDiff also uses spatial masks, it employs them in a different way.

Blended Diffusion \cite{blended-diffusion} and Blended Latent Diffusion \cite{blended-latent-diffusion} perform localized editing by blending the generated content from a text prompt with the noised version of the input image, using a user-provided mask in the pixel or latent space. 
DiffEdit \cite{diffedit} automatically derives masks from differences in noise predictions and blends content accordingly. PFB-Diff \cite{pfb-diff} progressively blends features across multiple layers of the U-Net and employs an attention masking strategy.

While both PFB-Diff and MINDiff apply masking in the cross-attention mechanism, they differ in where the mask is applied.
PFB-Diff applies a mask to the attention scores to restrict the spatial influence of specific textual tokens.
In contrast, MINDiff introduces an auxiliary attention branch, where a spatial mask is applied to its output.

Unlike prior methods~\cite{blended-diffusion, blended-latent-diffusion, diffedit, pfb-diff}, which primarily use masks to spatially blend reconstructed and edited content, MINDiff employs them to suppress subject influence in irrelevant regions through negative attention---a structurally distinct approach.

\subsection{Personalization}
\label{sec:personalization}

While image editing focuses on modifying specific parts of generated images, personalization enables a generative model to capture subject-specific characteristics from a few samples and generate new images that faithfully represent them in diverse contexts.

Textual inversion \cite{textual-inversion} learns the representation of a subject as a pseudoword embedding. However, its slow convergence and limited subject fidelity pose challenges to high-quality image generation. P+ \cite{voynov2023p+} and NeTI \cite{neti} extend Textual Inversion by dynamically generating learnable tokens at each denoising step or integrating U-Net layers to enrich object representations. DreamBooth \cite{dreambooth} utilizes an identifier token to represent a specific subject and updates the entire model weights to semantically entangle the identifier token with the specific subject. However, due to the full-model tuning strategy, it tends to overfit to the specific subject. To mitigate this, prior-preservation loss (PPL) is introduced during training by incorporating additional class-specific data, allowing the model to preserve general class characteristics. Custom Diffusion \cite{custom-diffusion} fine-tunes the key-value parameters of the cross-attention mechanism and also supports multi-concept learning.

\noindent\textbf{Overfitting mitigation.} Fine-tuning with limited samples often leads to overfitting. DisenBooth \cite{disenbooth} addresses this issue by disentangling the subject-relevant and subject-irrelevant embedding and minimizing their similarity. SID \cite{sid} reduces unwanted bias in subject embeddings by explicitly describing various elements of the reference images. This prevents non-subject features from being entangled with the subject representation. AttnDreamBooth \cite{pang2024attndreambooth} seperates the learning of embedding alignment, attention maps, and subject identity in order to better preserve subject's identity and reduce overfitting. Break-A-Scene \cite{break-a-scene} adopts a two-stage training scheme to balance subject identity preservation and editability. In the first stage, only the newly-added tokens are optimized with a high learning rate, while in the second stage, the model is fine-tuned with a lower learning rate.

\noindent\textbf{Controllability.} Recent methods have introduced user-controllable mechanisms that enable users to balance subject fidelity and text alignment.
COFT \cite{coft} proposes an optimization scheme constrained within a pre-defined radius around the pretrained model, enabling stable training but lacking post-training flexibility. In contrast, methods such as NeTI \cite{neti} and IP-Adapter \cite{ip-adapter} offer inference-time control. NeTI introduces importance-based ordering, allowing the user to adjust the trade-off between subject fidelity and text alignment by modifying a truncation threshold---higher thresholds increase subject fidelity. IP-Adapter~\cite{ip-adapter} introduces a decoupled cross-attention mechanism that enables adjustment of the weight assigned to image-based conditions.

\noindent\textbf{Cross-attention control.} Break-A-Scene \cite{break-a-scene} separates multiple subjects within a single image by learning to align each concept’s attention map with its corresponding segmentation mask. This alignment enables the model to distinguish subjects spatially and reduce interference between them. Cones2 \cite{cones2} receives a spatial layout as input and adjusts cross-attention activations to ensure accurate object positioning and reduce interference between objects.

\section{Method}
\label{sec:method}

\begin{figure}
    \centering
    \includegraphics[width=1.0\linewidth]{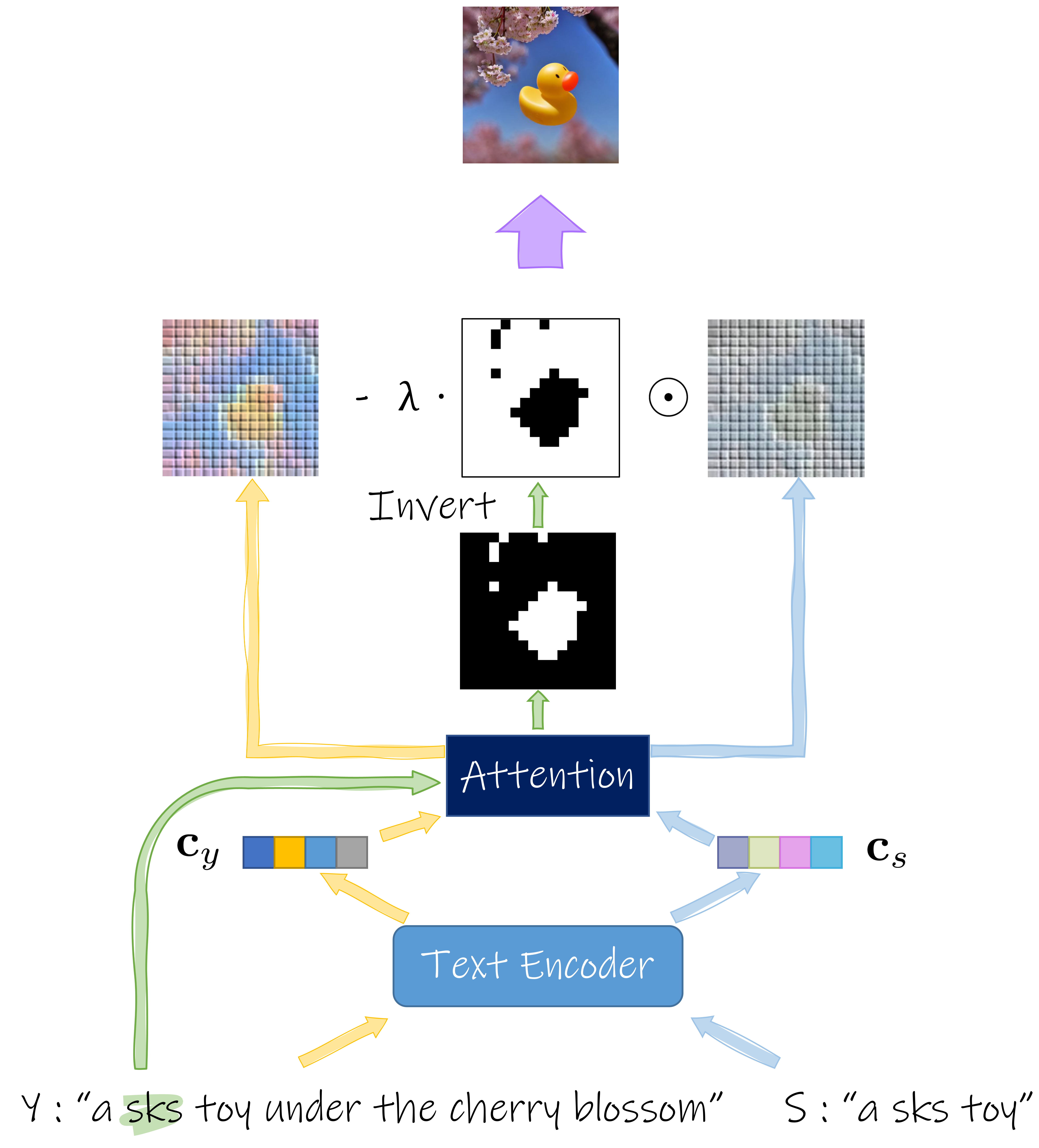}
    \caption{Illustration of the Mask-Integrated Negative Attention Diffusion (MINDiff) framework. The main text prompt $y$ and subject prompt $s$ are processed in separate attention branches to compute main attention and subject-specific auxiliary attention, respectively. A spatial mask, obtained by inverting the attention map of the identifier token, is applied to the auxiliary branch to localize suppression. The masked auxiliary attention is subtracted from the main attention. The suppression strength is controlled by a user-defined hyperparameter $\lambda$. This figure is conceptual; actual attention visualizations are provided in the supplementary material.}
    \label{fig:framework}
\end{figure}

\begin{figure*}[!t]
    \centering
    \includegraphics[width=1.0\linewidth]{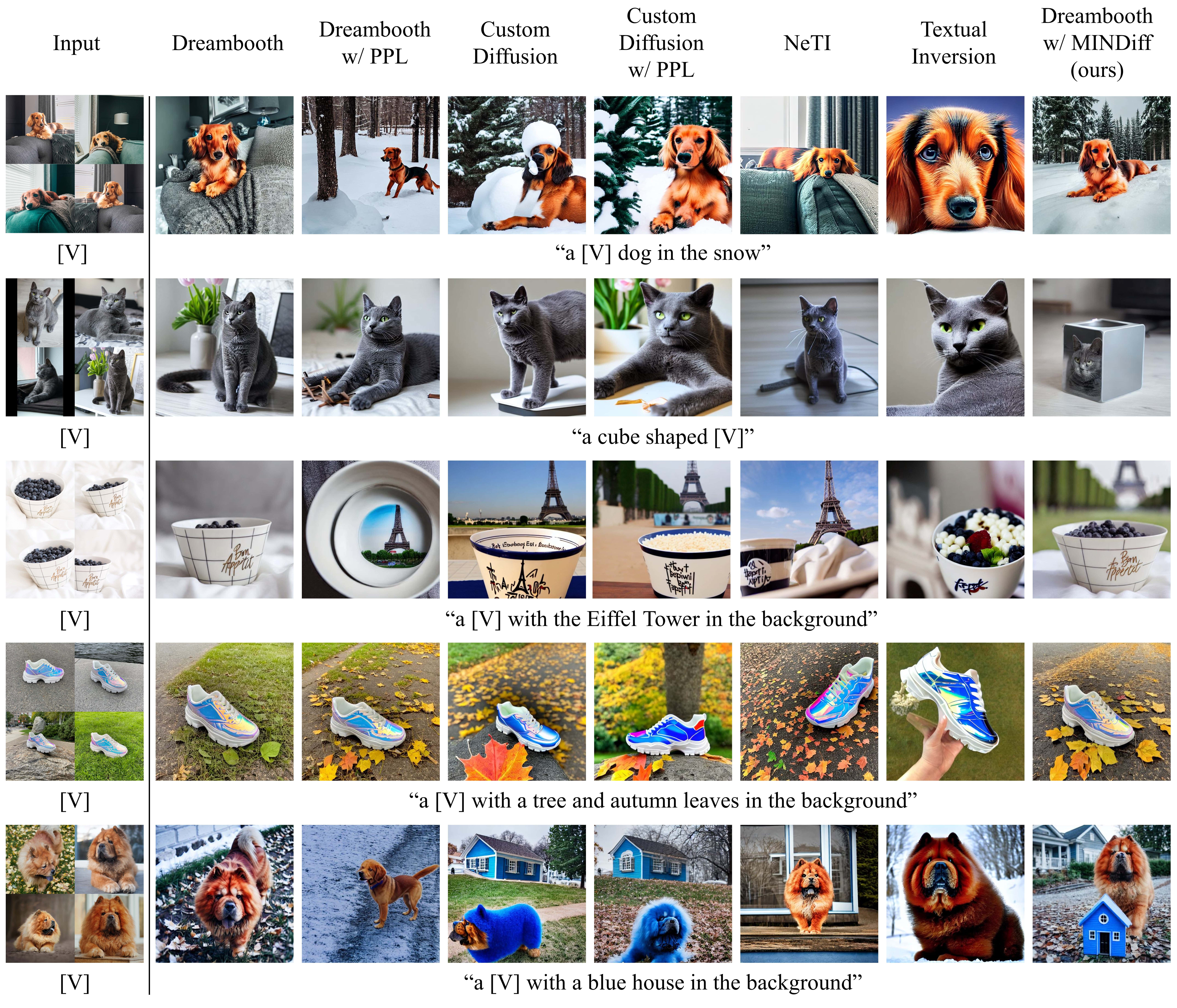}
    \caption{Qualitative comparison of text-conditioned image generation using Stable Diffusion 1.4 across personalization models: DreamBooth, DreamBooth with PPL, Custom Diffusion, Custom Diffusion with PPL, NeTI, Textual Inversion, and DreamBooth with MINDiff (ours). All images are generated using the same random seed for fair comparison. For NeTI, a truncation value of 64 is selected from the range (0, 128]. For MINDiff, a suppression scale of $\lambda = 0.6$ is used. [V] denotes the subject representation tokens.}
    \label{fig:compare_result}
\end{figure*}

\subsection{Latent Diffusion Models}
\label{sec:latent_diffusion_models}
Our approach builds upon latent diffusion models (LDMs) \cite{ldm}, where high-dimensional input data $\mathbf{x}_0$ is projected into a latent space representation $\mathbf{z}_0$ using an encoder $\mathcal{E}(\cdot)$.
The forward process of LDMs is formulated as a Markov chain, where random Gaussian noise ${\boldsymbol{\epsilon}}$ is gradually added to the latent vector over $T$ steps. Mathematically, this process is expressed as follows:
\begin{equation}
\mathbf{z}_t=\sqrt{\bar\alpha_t}\mathbf{z}_0+\sqrt{1-\bar\alpha_t}\boldsymbol{\epsilon},
\quad
\boldsymbol{\epsilon}\sim\mathcal{N}(\mathbf{0},\mathbf{I}),
\label{eq:forward_process}
\end{equation}
where $\alpha_t$ is a coefficient that controls the noise variance, and its cumulative form is given by $\bar{\alpha_t}=\prod_{s=1}^t\alpha_s$.

The reverse process aims to reconstruct the original latent representation $\mathbf{z}_0$
by de-noising progressively added noise from the forward process. Since the Markov process used in diffusion models is not inherently reversible, the model is trained to estimate the noise or the reverse transition probabilities at each time step.
To incorporate additional conditioning information, we use a conditional embedding $\mathbf{c}$, and the model is trained to minimize the loss of mean squared error (MSE) between the predicted noise and ground-truth noise.
\begin{equation}
    \mathbb{E}_{\mathbf{z}\sim\mathcal{E}(\mathbf{x}),\boldsymbol{\epsilon} \sim \mathcal{N}(\mathbf{0},\mathbf{I}), t}\left[\|\boldsymbol{\epsilon} - \boldsymbol{\epsilon}_\theta(\mathbf{z}_t, t,\mathbf{c})\|^2\right].
\label{eq:loss}
\end{equation}

After training, the model generates final images by initializing from a random latent vector $\mathbf{z}_T \sim \mathcal{N}(\mathbf{0}, \mathbf{I})$ in the latent space and iteratively applying the reverse process. The reconstructed latent representation $\mathbf{z}_0$ is then passed through a decoder $D(\cdot)$ yielding the final output image $\mathbf{x}_0$.

\subsection{Mask-Integrated Negative Attention Diffusion (MINDiff)}
\label{sec:mindiff}

The proposed Mask-Integrated Negative Attention Diffusion (MINDiff) is designed to mitigate the overfitting problem commonly observed in optimization-based personalization models such as DreamBooth. MINDiff assumes access to a personalization model that has already been fine-tuned on a target subject and operates entirely at inference time, without requiring any additional optimization.

To achieve this, MINDiff modifies the cross-attention mechanism~\cite{vaswani2017attention} by introducing an auxiliary attention branch, whose output is assigned a negative weight—forming what we refer to as negative attention.
This negative attention suppresses the subject’s influence in spatially masked regions and enables more balanced image generation.
This design is conceptually distinct from the standard cross-attention mechanism used in diffusion models.
To the best of our knowledge, this is the first method to incorporate negatively weighted attention in a spatially controlled manner.
The modified cross-attention mechanism is formulated as follows:

\begin{equation}
\begin{aligned}
    Z &= \text{Softmax}\left(\frac{Q K^\top}{\sqrt{d_k}}\right)V\\
    &\quad\quad\quad\quad\quad\quad - \lambda  \cdot \text{Mask} \circ \text{Softmax}\left(\frac{Q K_s^\top}{\sqrt{d_k}}\right) V_s,
\end{aligned}
\end{equation}
where $\lambda$ serves as a scaling factor that regulates the intensity of negative attention; a higher value of 
$\lambda$ results in a stronger suppression of the influence of subject. Here, $d_k$ denotes the dimensionality of the key and query vectors, which are assumed to share the same embedding space.

The input prompt $y$ specifies the main text condition, while an additional subject prompt $s$ is provided to indicate the personalized subject, typically containing an identifier token (e.g., “sks”).
After passing through the text encoder, the prompt $y$ is transformed into the conditioning vector $\mathbf{c}_y$ and the subject prompt $s$ is transformed into the conditioning vector $\mathbf{c}_s$. The latent image feature $\mathbf{f}$ is projected to the Query matrix.
The vectors $\mathbf{c}_y$ and $\mathbf{c}_s$ are then used to compute the Key and Value matrices for the main and auxiliary attention branches, respectively, with the latter producing the negative attention term.
These projections are computed as follows:

\begin{equation}
\begin{aligned}
    Q = \ell_Q(\mathbf{f}), \quad
    K &= \ell_K(\mathbf{c}_y), \quad
    V = \ell_V(\mathbf{c}_y) \\
    K_s &= \ell_K(\mathbf{c}_s), \quad V_s= \ell_V(\mathbf{c}_s),
\end{aligned}  
\end{equation}

where $\ell_Q,\ell_K,\ell_V$ are learned linear projections.


As illustrated in \cref{fig:framework}, the auxiliary attention branch takes as input the subject prompt $s$ and produces a suppressive term, referred to as negative attention. To determine where this negative attention should be applied, we use a spatial mask that identifies non-subject (background) regions. This mask is generated at each denoising step using the attention map extracted from the identifier token.

We follow the technique in MasaCtrl~\cite{cao2023masactrl}, where an attention map at a 16×16 resolution is computed across all attention heads and layers at the same time step and then averaged. While MasaCtrl~\cite{cao2023masactrl} applies this setting to Stable Diffusion 1.4, we adopt the same setting for Stable Diffusion 1.4 (SD1.4) and extend it to other versions by using 24×24 for for Stable Diffusion 2.1 (SD 2.1) and 32×32 for Stable Diffusion XL (SDXL). The resulting averaged attention map is binarized by assigning a value of 1 to elements above the mean, producing a subject mask. 
We then invert this mask to isolate background regions. This effectively prevents the model from becoming unnecessarily similar to the reference image and ensures more faithful alignment with the given text prompt.


\section{Experiments}
\label{sec:experiments}

\begin{figure*}[!t]
    \centering
    \includegraphics[width=1.0\linewidth]{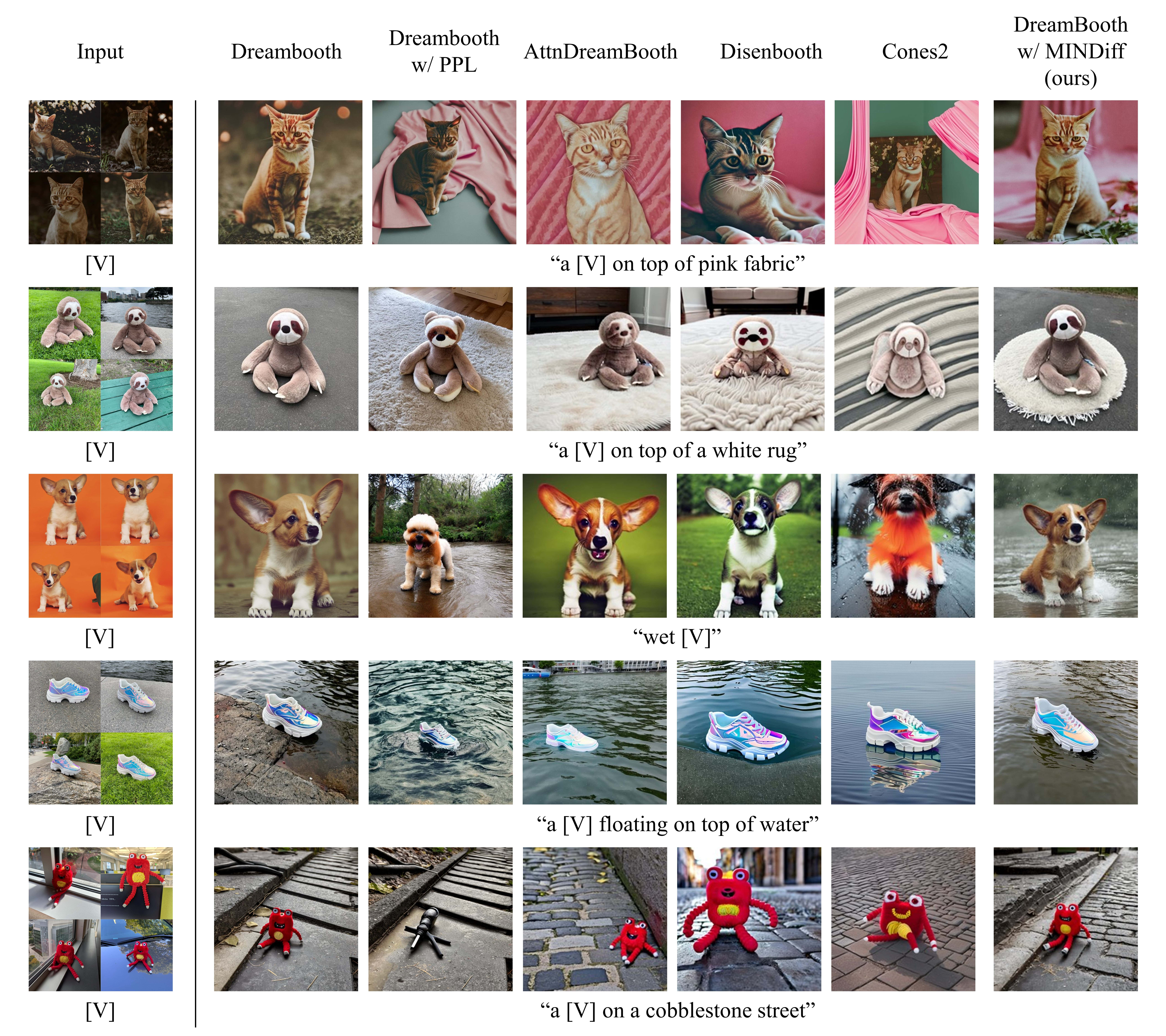}
    \caption{Qualitative comparison of text-conditioned image generation using Stable Diffusion 2.1 across personalization models: DreamBooth, DreamBooth with PPL, AttnDreamBooth, DisenBooth, Cones2, and DreamBooth with MINDiff (ours). All images are generated using the same seed for fair comparison. For MINDiff, a suppression scale of $\lambda = 0.6$ is used.}
    \label{fig:comparison_sd21}
\end{figure*}

\begin{figure}
    \centering
    \includegraphics[width=1.0\linewidth]{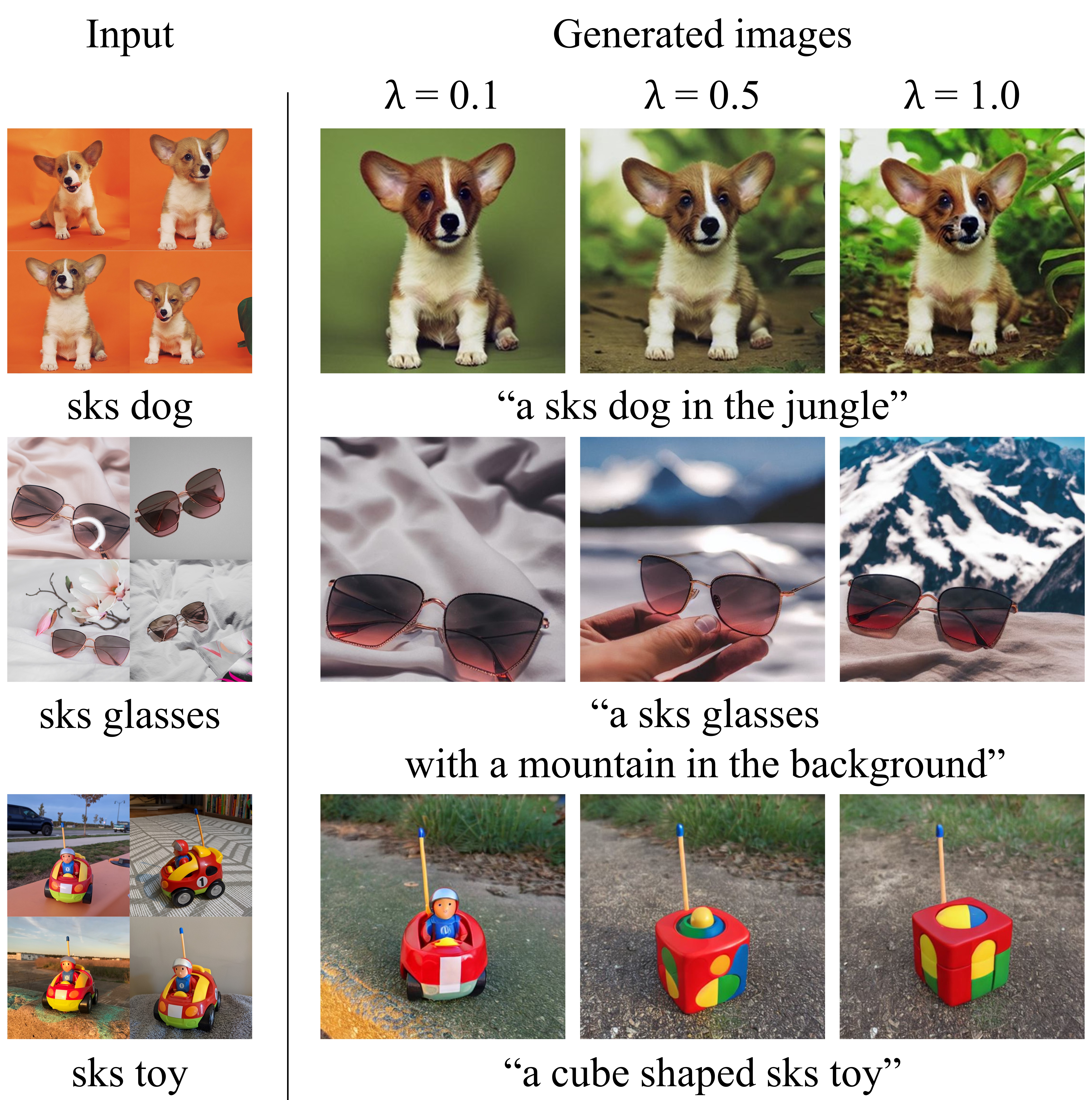}
    \caption{Qualitative results of DreamBooth with MINDiff using different values of the scale parameter $\lambda$. Each row is generated with the same seed. As $\lambda$ increases, subject influence is suppressed, allowing the model to better reflect the text prompt.}
    \label{fig:scale_comparison}
\end{figure}

\begin{figure*}
    \centering
    \includegraphics[width=1.0\linewidth]{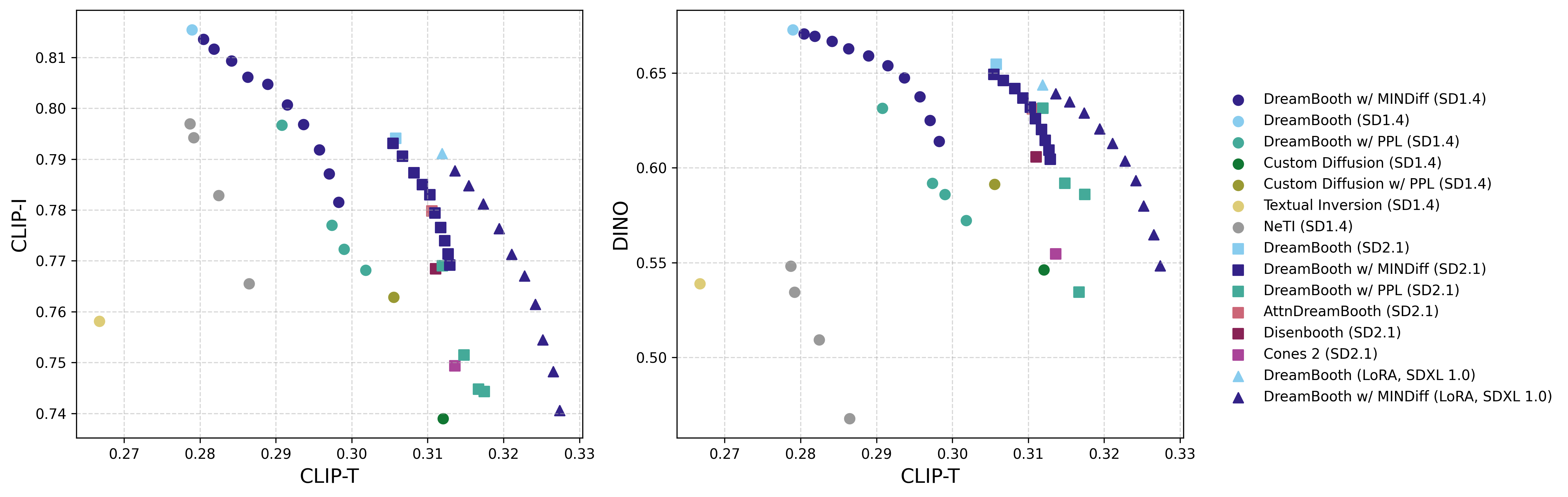}
    \caption{Quantitative comparison of text-conditioned image generation using CLIP-T similarity (x-axis) against two image-based metrics: CLIP-I (left panel, y-axis) and DINO (right panel, y-axis). All metrics evaluate the alignment between generated images and either the text prompt (CLIP-T) or the reference image (CLIP-I, DINO). Marker shapes indicate different Stable Diffusion versions (circle: SD 1.4, square: SD 2.1, triangle: SDXL 1.0), and personalization methods are color-coded as shown in the legend.}
    \label{fig:clip_result}
\end{figure*}

\subsection{Experimental Setup}
\label{sec:experimental_setup}

We evaluate our method, MINDiff, across multiple Stable Diffusion backbones. For SD 1.4, we compare with widely used personalization methods, including Textual Inversion~\cite{textual-inversion}, DreamBooth~\cite{dreambooth}, Custom Diffusion~\cite{custom-diffusion}, and NeTI~\cite{neti}. For SD 2.1, we compare with more recent methods such as AttnDreamBooth~\cite{pang2024attndreambooth}, DisenBooth~\cite{disenbooth}, and Cones2~\cite{cones2}. Finally, to assess the applicability of MINDiff to SDXL, we apply it to a LoRA-based DreamBooth comparing results with and without our method.

For both training and evaluation, we employ the DreamBooth Dataset \cite{dreambooth}.
We denote the representation for each subject as [V], following the instructions of each method (\eg, ``sks dog'', ``dog'', or other textual representations).

\subsection{Qualitative Results}
\label{sec:qualitative_results}

\Cref{fig:mindiff_result} demonstrates that DreamBooth with MINDiff effectively reflects the text prompt while preserving subject identity. It also enhances diversity and visual quality across generations.

As shown in \cref{fig:compare_result,fig:comparison_sd21}, existing methods often struggle to balance subject identity preservation and editability, and may even generate incorrect subjects. In contrast, MINDiff consistently achieves a more balanced trade-off between text alignment and subject fidelity across diverse prompts and environments. We also evaluate our method on SDXL and observe similar improvements.

\Cref{fig:scale_comparison} presents a comparison of generated images from DreamBooth with MINDiff across different $\lambda$ values.
By adjusting $\lambda$, users can generate images tailored to their specific goals or intentions.
As $\lambda$ increases, the generated images align more closely with the given text conditions.

\subsection{Quantitative Results}
\label{sec:quantitative_results}

\Cref{fig:clip_result} presents a quantitative evaluation using both CLIP \cite{clip} and DINO \cite{dino} based similarity metrics. For text alignment, we use CLIP-T, which computes the cosine similarity between the CLIP embeddings of the text prompt and the generated image. For subject fidelity, we employ two metrics: CLIP-I, which measures the similarity between the generated image and the reference image, and the DINO metric, proposed by DreamBooth~\cite{dreambooth}, which averages pairwise cosine similarity between DINO ViT-S/16 embeddings. Unlike CLIP, which is trained with supervision on image-text pairs, DINO is trained in a self-supervised manner to capture unique subject-specific features, making it more suitable for evaluating identity preservation.

We sweep PPL weights (loss-term scales of 0.1, 0.5, 0.75, and 1.0) and MINDiff scales (negative-attention strengths ranging from 0.1 to 1.0). On the SD 1.4 backbone, at specific PPL weights (0.1 and 0.5), MINDiff achieves higher scores in both text alignment and subject fidelity for certain scale ranges. This demonstrates that MINDiff is more effective than PPL in mitigating overfitting in personalization.
DreamBooth with PPL requires class-specific data—resulting in longer training time—and a fixed loss weight that must be set in advance, which reduces flexibility.
In contrast, MINDiff introduces a tunable scale parameter $\lambda$ that can be adjusted at inference time, allowing users to steer generation results without retraining.
Finally, compared to NeTI---a controllable baseline---DreamBooth with MINDiff consistently outperforms across all evaluated metrics.

On the SD 2.1 backbone, MINDiff offers comparable control to its behavior on SD 1.4. While it does not show as clear an improvement over prior-preservation loss as in SD 1.4, it retains the key advantage of inference-time controllability, which remains valuable in practical applications.

Finally, MINDiff also works effectively with LoRA-based DreamBooth on the SDXL backbone. These results demonstrate that MINDiff consistently works across multiple versions of Stable Diffusion, including SD 1.4, SD 2.1, and SDXL.

\subsection{Ablation Study}
\label{sec:ablation_study}

\begin{figure}[htbp]
    \centering
    \includegraphics[width=1.0\linewidth]{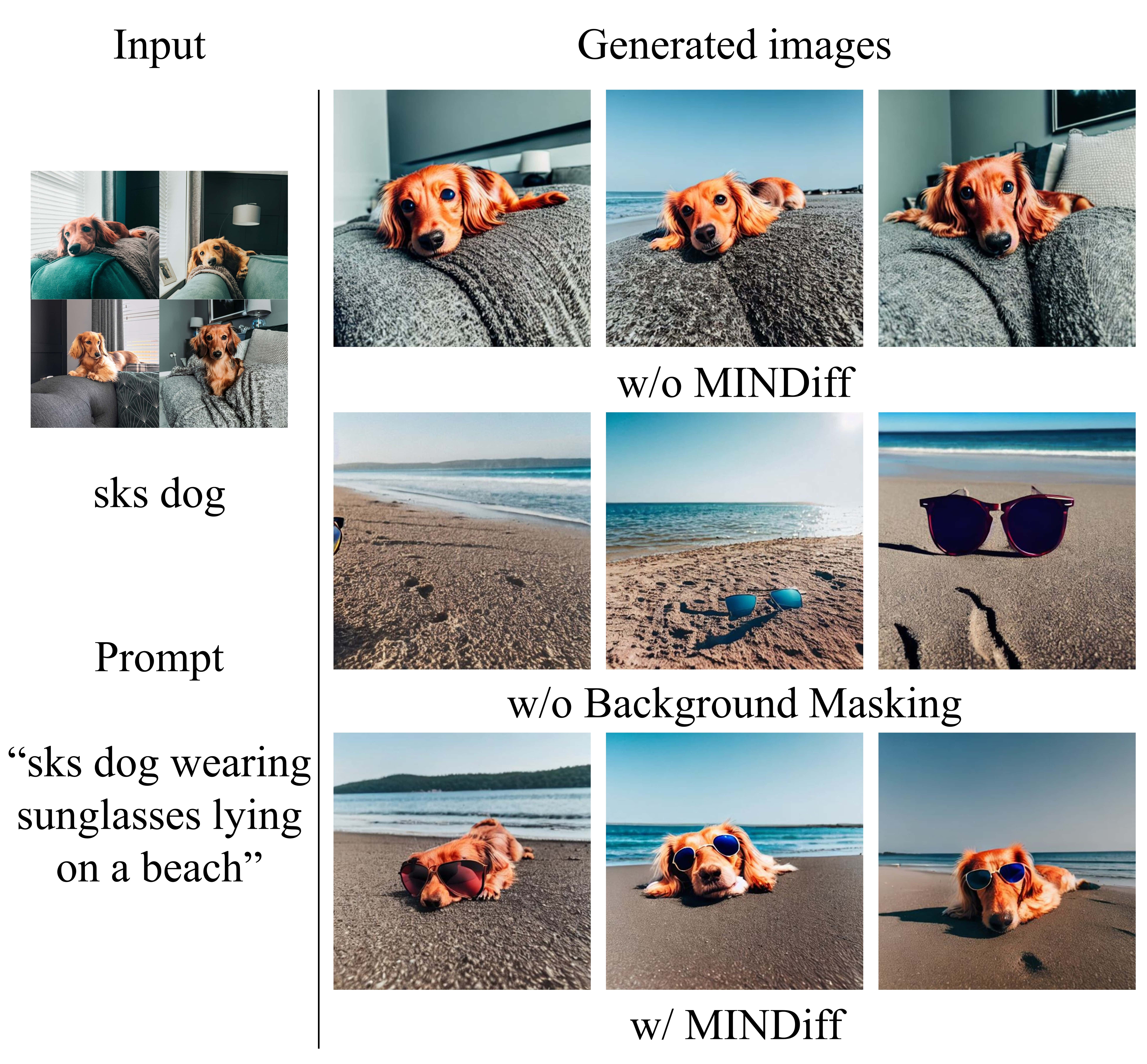}
    \caption{Ablation study on the effect of Background Masking in MINDiff. Each column is generated using the same seed. The first row shows results from DreamBooth without MINDiff. The second row applies negative attention without background masking. The third row presents the full MINDiff method with both negative attention and background masking applied. \texttt{w/o Background Masking} denotes the setting where negative attention is used without the spatial mask.}
    \label{fig:ablation_study}
\end{figure}

The previous study \cite{dreambooth} suggested that fine-tuning with DreamBooth could degrade the ability of the model to generate diverse outputs, leading to the introduction of prior-preservation loss. However, our experiments reveal that this degradation does not occur inherently. Instead, we identify that strong influence from the learned identifier token weakens the effect of the text prompt, leading to prompt dilution.

\Cref{fig:ablation_study} presents an ablation study on Background Masking. \texttt{w/o Background Masking} refers to applying negative attention without the mask. When Background Masking is not applied, the subject (\eg,``a sks dog'') is completely removed, while only the remaining portion of the prompt (``wearing sunglasses lying on a beach'') is reflected. This indicates that negative attention effectively controls text influence. However, if its influence is not properly constrained, it may lead to subject loss. These findings highlight the importance of Background Masking when applying negative attention.
\section{Limitations}
\label{sec:limitations}

The proposed MINDiff method mitigates overfitting without prior-preservation loss and enables controllable generation at inference time. However, it has several limitations.

First, the model may still suffer from language drift due to strong entanglement between subject and class token, making it difficult to fully disentangle these two components. For instance, when a model is trained on the prompt ``a photo of a sks dog'', it often generates the learned subject even when only ``dog'' is provided at inference time.

Second, our method relies on a manually tuned scale parameter $\lambda$ to control the trade-off between identity preservation and editability. While this provides flexibility at inference time, it also introduces a usability challenge: inappropriate values of $\lambda$ may lead to generation failures, such as missing the subject or failing to follow the prompt. In our experiments, we found that $\lambda = 0.6$ often provides a good balance across diverse prompts and subjects. However, automating the tuning of this parameter could further improve the usability. We leave this as an avenue for future work.

\section{Conclusion}
\label{sec:conclusion}

We presented MINDiff, a novel inference-time method for mitigating overfitting in personalized text-to-image generation. Our key contribution is the introduction of negative attention, which suppresses subject influence in irrelevant regions via a spatial mask. This approach extends the standard cross-attention mechanism without requiring any changes to model weights.

MINDiff effectively mitigates overfitting without relying on class-specific prior-preservation loss, avoiding additional training cost. It also enables users to control the trade-off between subject fidelity and text alignment during inference. Extensive experiments demonstrate that MINDiff mitigates overfitting more effectively than DreamBooth with PPL, while outperforming NeTI, a controllable baseline, in both subject fidelity and text alignment.

Since MINDiff operates entirely at inference time, it can be directly applied to existing fine-tuned DreamBooth models, making it a practical and flexible solution for real-world applications.

{
    \small
    \bibliographystyle{ieeenat_fullname}
    \bibliography{main}
}

\end{document}